\documentclass[letterpaper, 10 pt, conference]{ieeeconf}  

\IEEEoverridecommandlockouts                              

\usepackage{cite}
\usepackage{amsmath,amssymb,amsfonts}

\usepackage{graphicx}
\usepackage{textcomp}
\usepackage{xcolor}
\usepackage{subcaption}
\usepackage{algorithmicx}
\usepackage{algorithm}
\usepackage{algpseudocode}

\title{\LARGE \bf
LocoNeRF: A NeRF-based Approach for Local Structure from Motion for Precise Localization
}

\author{Artem Nenashev, Mikhail Kurenkov, Andrei Potapov, Iana Zhura, Maksim Katerishich, and Dzmitry Tsetserukou
\thanks{The authors are with the Intelligent Space Robotics Laboratory, Skoltech, Bolshoy Boulevard 30, bld. 1, 121205, Moscow, Russia }
\thanks{Email: \{Artem.Nenashev, Mikhail.Kurenkov, Andrei.Potapov, Iana.Zhura, Maksim.Katerishich, D.Tsetserukou\}@skoltech.ru
}
}

\begin{document}

\maketitle
\thispagestyle{empty}
\pagestyle{empty}

\maketitle

\begin{abstract}
Visual localization is a critical task in mobile robotics, and researchers are continuously developing new approaches to enhance its efficiency. In this article, we propose a novel approach to improve the accuracy of visual localization using Structure from Motion (SfM) techniques. We highlight the limitations of global SfM, which suffers from high latency, and the challenges of local SfM, which requires large image databases for accurate reconstruction. To address these issues, we propose utilizing Neural Radiance Fields (NeRF), as opposed to image databases, to cut down on the space required for storage. We suggest that sampling reference images around the prior query position can lead to further improvements. We evaluate the accuracy of our proposed method against ground truth obtained using LIDAR and Advanced Lidar Odometry and Mapping in Real-time (A-LOAM), and compare its storage usage against local SfM with COLMAP in the conducted experiments. Our proposed method achieves an accuracy of 0.068 meters compared to the ground truth, which is slightly lower than the most advanced method COLMAP, which has an accuracy of 0.022 meters. However, the size of the database required for COLMAP is 400 megabytes, whereas the size of our NeRF model is only 160 megabytes. Finally, we perform an ablation study to assess the impact of using reference images from the NeRF reconstruction.

\end{abstract}

\section{Introduction}

\subsection{Motivation}
\begin{figure}[!htb]
\centering
\includegraphics[width=0.48\textwidth]{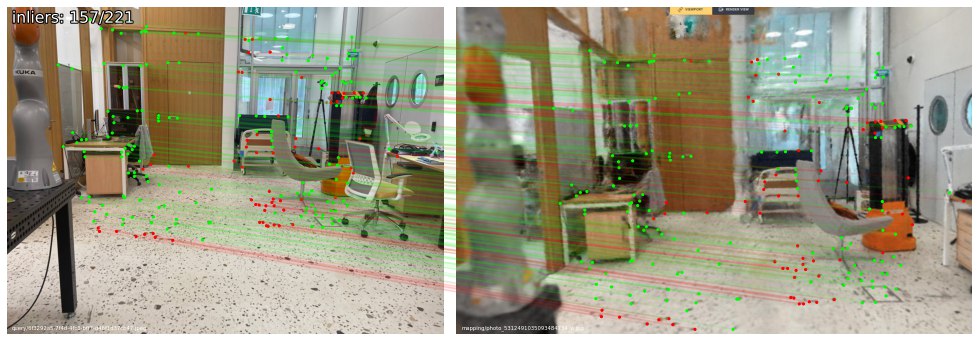}
\caption{Feature matching of query image and reference image rendered from neural radians field map, where green points are the key points and red points is consensus points that are used to refine the estimated pose of the camera}
\label{fig:fig2}
\end{figure} 

Localization is a key task in mobile robotics, and the solution of which is necessary for navigation and control tasks, such as logistics robots. The development of existing localization methods and the creation of new approaches have gained popularity over the years. The rise of modern machine learning and deep learning technologies in computer vision has also made it possible to create approaches that can compete with classical methods \cite{chen2020survey}. Solving this problem can reduce the cost of robots, the time required for robot travel, and the number of incidents involving people\cite{kalinov2020warevision, perminov2021ultrabot, karpyshev2022mucaslam, mikhailovskiy2021ultrabot}.


\subsection{Problem statement}
The existing approaches for solving visual localization suffer from low accuracy and are not real-time. To estimate the position accurately, most of the methods use Structure from Motion \cite{schonberger2016structure} which requires a large database of images and several days of computing time. 
Some SfM-based approaches suggest compressing the database to reduce memory costs for storage \cite{van2018efficient, mera2020efficient}, adapting localization algorithms to speed up work, or hybrid methods \cite{camposeco2019hybrid} that combine some advantages. However, there is still no approach that completely eliminates all the current shortcomings of localization methods.
Thus, the question of reducing memory costs, shortening time, and improving the accuracy and robustness of localization for mobile robots remains open.


\subsection{Related Works}
\textbf{Visual Localization.} Classical approaches to visual localization typically involve the map creation followed by localization on that map. SfM \cite{ullman1979interpretation} is one of the most popular and widely used methods for obtaining a 3D model of a scene from a set of 2D images and subsequent localization of query images. It allows for the use of data obtained from different cameras as well as additional sources of information, such as depth sensor data. However, SfM has the key drawbacks, including high data requirements for reconstruction and sensitivity to the selection of key points and changes in lighting when creating the scene model.

The Active Search method \cite{sattler2017large} proposes reformulating the task of Structure-from-Motion (SfM) localization by using a relatively small set of 3D landmarks, whose visibility and discriminability within the scene are carefully chosen to facilitate accurate pose estimation. The method is shown to provide comparable, or even better accuracy than existing approaches that use large-scale 3D models, while requiring substantially fewer computational resources. Additionally, it is robust to lighting and viewpoint changes, making it a suitable choice for real-world applications. The downside of this method is an increase in localization time when using 3D data as well as a lower accuracy when using only 2D data.


The hierarchical-localization (hloc)\cite{sarlin2019coarse} approach proposed an improvement and extension to the SfM method to enhance the reliability and robustness of localization. Standard handcrafted feature detectors (ORB\cite{rublee2011orb}, SURF\cite{bay2006surf}, SIFT\cite{ng2003sift}) were replaced with a monolithic network, HF-Net, which performs three important functions: extracting global descriptors using NetVLAD\cite{arandjelovic2016netvlad}, extracting local features using SuperPoint\cite{detone2018superpoint}, and estimating quality of local features. The use of global descriptors allowed for a hierarchical approach to localization and accelerated the method, while the NN-based approach to extracting local features resulted in improved robustness of localization.

The PixLoc method \cite{sarlin2021back} is an advanced method of hloc. It is based on visual localization of a query image in a known 3D scene by aligning it with reference images. The alignment is obtained by minimizing an error over deep features generated by a CNN. The CNN and optimization parameters are trained end-to-end from ground truth poses. The method utilizes a 3D representation of the environment, such as a sparse or dense 3D point cloud, and performs image alignment over learned feature representations of the images. Direct alignment is used to find the pose which minimizes the difference in appearance between the query image and each reference image. The CNN predicts an uncertainty map along with each feature map, which is used to steer the optimization towards the correct pose. The method learns to ignore dynamic objects and repeated patterns, and focuses on road markings, silhouettes of trees, or prominent structures on buildings. The optimizer is fitted to the distribution of poses or residuals but not to their semantic content, and the damping parameters vary with the training data. The method shows promising results in terms of accuracy and robustness to illumination and viewpoint changes.

The alternative to this is hybrid usage of data from LiDAR and camera sensors, presented by Yudin et al\cite{yudin2022cloudvision}. This is a novel method for visual localization that employs deep neural networks DNNs in combination with pre-built LIDAR point cloud data. The main idea behind this technique is to create a database of images and point cloud information, which can be used to estimate the robot's pose in real-time.

The algorithm utilizes image retrieval to find the most similar image in the database to the current sensor image for localization. The 3D point cloud data is used to refine the pose estimation and increase the accuracy of the localization. The pre-build LiDAR map allows for more robust and accurate localization, even in challenging environments, while DNNs for feature matching and retrieval allows for efficient and accurate image alignment. However one of the main limitations is the need for an existing database of pre-captured images and point cloud data, which can be time-consuming and may not be suitable for dynamic environments. 



As an alternative to SfM, NeRF\cite{mildenhall2021nerf} also allows for creating photo-realistic 3D scene reconstructions using only 2D images. The key feature of NERF is representing space in an implicit form using MLP. This approach allows for representing the light transport function in the environment as weights, creating a continuous representation of density and color at each point of the reconstructed space. The basic version of NERF required days of training, but the results were stable only for small scenes where the images covered a large part of the space from different angles. The approach was subsequently expanded using new techniques to improve robustness and reconstruction quality.

\begin{figure}[!htb]
\centering
    \includegraphics[width=0.4\textwidth]{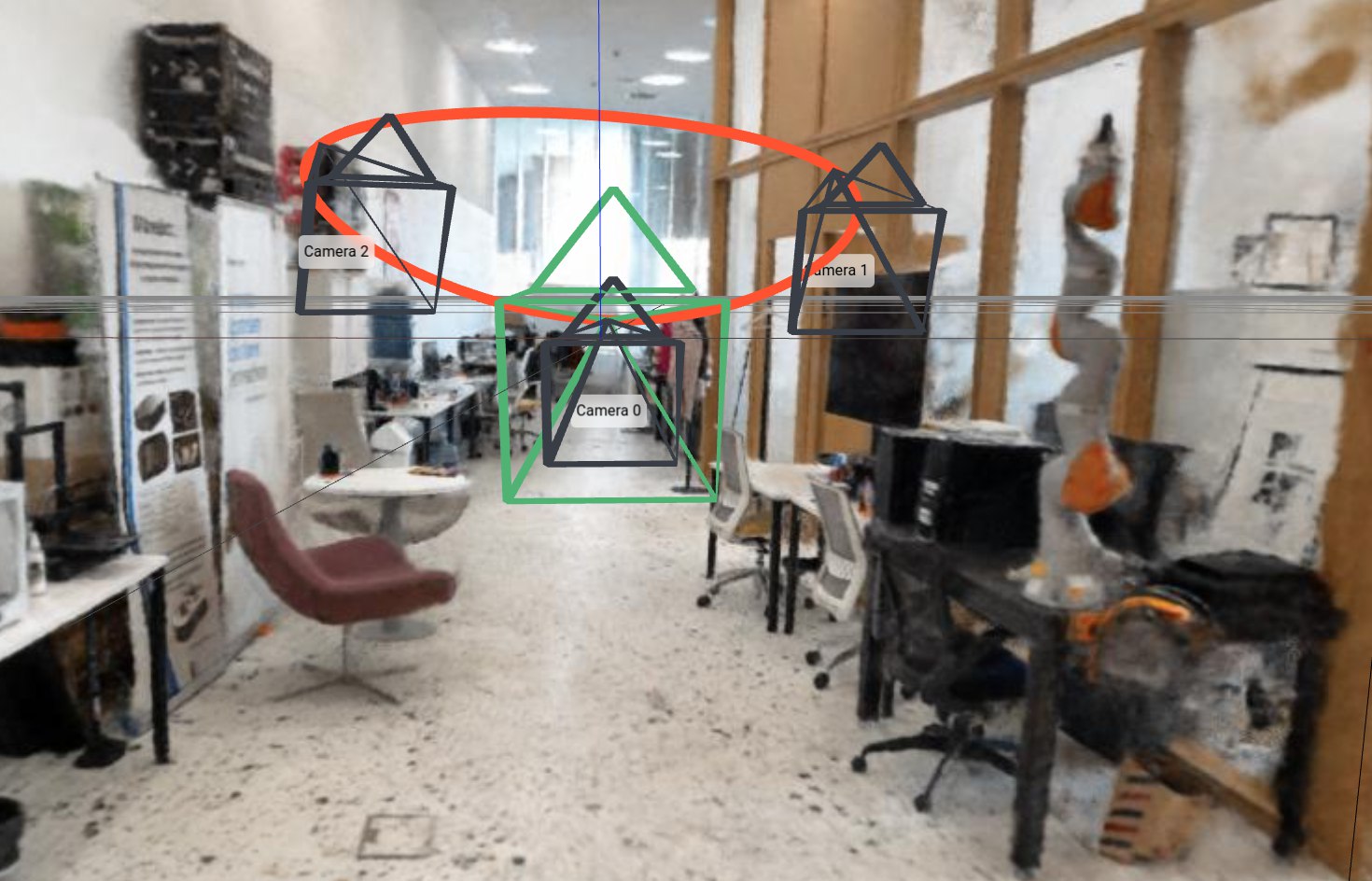}
    \caption{Image extraction from NeRF. Camera 0 is the query image, Camera 1 and Camera 2 are the reference images, which were used for localization of query.} \label{fig:method-nerf}
\end{figure}

Thus, the architecture proposed in Mip-NeRF \cite{barron2021mip} used conical frustums instead of rays to reduce aliasing artifacts and improve the network's ability to represent fine details. This work was further developed in Mip-NeRF 360\cite{barron2022mip}, where a number of improvements were made to enhance the rendering of unbounded scenes, including non-linear scene parametrization, online distillation, and a novel distortion-based regularization. One of the method expansion was suggested in NERF++\cite{zhang2020nerf++}, where the parametrization of unbounded scenes was also improved.

 There have been also introduced several improvements to the encoding methods used in NeRF. For instance, Mip-NeRF incorporates integrated positional encoding (IPE) which enables encoding of Gaussian distributions describing conical frustums. This allows for more accurate representations of complex scenes. In another work, SIREN\cite{sitzmann2020implicit} encoding was proposed as a replacement for PE to enhance the handling of complex signals and increase robustness. By learning the encoding, the network is able to better adapt to the input data and improve performance.

Instant-NGP\cite{muller2022instant} presented multiresolutional hash encoding which stores input parameters as a set of feature vectors in hash tables, indexed based on resolution and concatenated before being fed into the neural network. This approach not only achieves high-quality reconstructions but also reduces training time to a few tens of minutes per scene. This is a significant improvement over previous methods that required hours or even days to train on a single scene.

Currently, one of the most modern solutions that combines the advantages of previous approaches is Nerfacto\cite{tancik2023nerfstudio}. It is a modular framework that allows experimentation with different combinations of existing solutions and a separate method that incorporates successful architectural solutions from other approaches. Nerfacto achieves high-quality photo-realistic reconstructions while minimizing the time required for their generation. By combining the best aspects of previous methods, this approach is a significant step forward in the field of neural rendering.

\begin{figure}[h!]
\centering
    \includegraphics[width=0.4\textwidth]{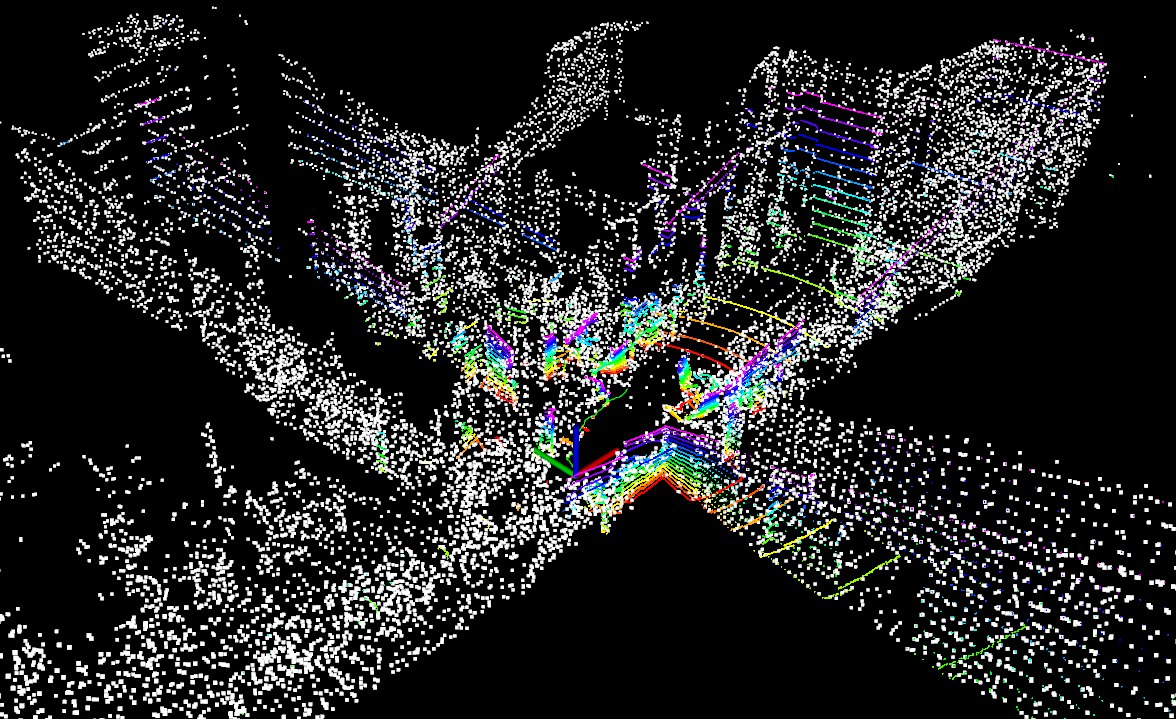}
    \caption{Ground truth trajectory estimation with A-LOAM method.} 
    \label{fig:point_cloud}
\end{figure}

For localization in Loc-NeRF \cite{maggio2022loc}, a Monte Carlo algorithm was proposed using NERF as a map. In this case, particles are represented by 6-DoF camera poses, and the likelihood is computed based on the similarity between the observed image and the rendered NeRF image. This approach improves robustness and localization quality, but a significant challenge is the time required to create the map and render images using NeRF. Despite this challenge, the approach shows promise and has the potential to be a valuable tool in localization and mapping applications.
The introduced method utilizes the local SfM approach, which allows for accurate camera pose estimation by leveraging visual features in the local neighborhood. However, to further enhance the accuracy and robustness of the method, photorealistic NeRF renders are added. This improves the localization of the camera and increases the robustness compared to SfM methods alone. Additionally, the incorporation of NeRF renders helps to mitigate some of the issues associated with NeRF, such as the time required for rendering and the sensitivity to lighting conditions. Overall, the combination of local SfM and Nerf renders is a promising approach for accurate and robust camera localization in complex scenes.

\subsection{Contribution}
The goal of this research is to investigate the potential benefits of NeRF implementation as a proxy for image databases in Local SfM techniques. The primary focus is on reducing the latency and storage requirements of the method while maintaining or improving the accuracy of the results. Additionally, the research aims to explore the possibility of further accuracy improvements of the method by incorporating sampling for reference images around the prior query position.
\textbf{The main contributions of this work are:}
\begin{itemize}
    \item development of a Local SfM method that uses NERF instead of image databases as a proxy for reference images;
    \item evaluation of the efficiency and accuracy of the NERF-based method compared to traditional Local SfM techniques that use image databases;
\end{itemize}

\begin{figure*}[h]
\centering
\includegraphics[width=0.9\textwidth]{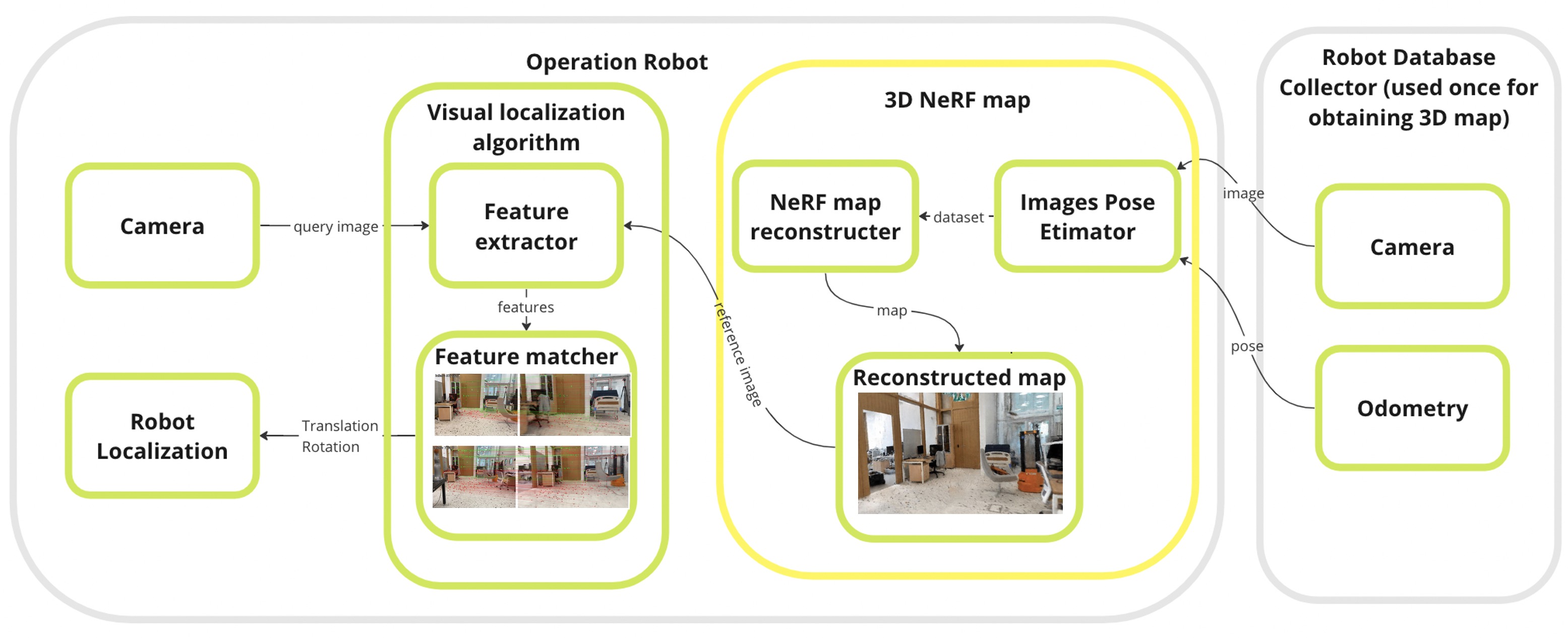}
\caption{Proposed visual localization based on Structure From Motion with Neural Radiance Fields map pipeline.}
\label{fig:main_schema}
\end{figure*} 
\section{Method Overview}

Proposed method for visual localization can be divided into three main parts. The first step involves obtaining data using a LiDAR-based SLAM algorithm to create a NeRF reconstruction of the environment. The second step involves training a DNN to create a implicit representation of the environment in the weights of NeRF. Finally, the third step involves using the rendered images from NeRF at the current position, which are then matched with the query image obtained from the robot camera.
Method leverages the strengths of both LiDAR-based SLAM algorithms and DNNs to create an accurate and robust 3D reconstruction of the environment, which can then be used for visual localization. The use of NeRF reconstruction enables the rendering of high-quality images at the current position, improving the accuracy of image matching with the query image.

To create the NeRF reconstruction of the environment, we needed to obtain images with their corresponding positions in the environment. We used a LiDAR-based SLAM algorithm to obtain the images and positions. After analyzing the available SLAM algorithms, the LOAM \cite{zhang2014loam} \cite{wang2021f} algorithm has been implemented due to its leading performance in benchmarks. The result of a point cloud is provided in Fig.\ref{fig:point_cloud}.

To achieve more accurate and robust results, we employed the advanced version of LOAM, known as A-LOAM, which facilitated the acquisition of high-quality images of the environment with their corresponding positions. The images were subsequently preprocessed to eliminate any blurry images and ensure that only the highest quality images were utilized in the NeRF reconstruction process.

Once the images and corresponding positions from different perspectives were obtained, the NeRF model was trained to generate a 3D reconstruction of the environment Fig.\ref{fig:method-nerf}. To achieve this, we utilized the nerfacto method within the NeRF Studio tool. NeRF Studio is a user-friendly tool that enables users to train and render NeRF models effortlessly.

The localization algorithm consists of several steps. Firstly, the input image is acquired. Next, four images are rendered from NeRF based on the previous position, located near the prior coordinates. Descriptors are then extracted from all images, and the SFM method is used to obtain the rotation and translation to the input image. Improved accuracy of the method requires multiple reference images. Full pipeline is presented in Fig. \ref{fig:main_schema}.

\section{System Overview}
The mobile platform employed for collecting the dataset and executing the algorithm was the HermesBot, a cutting-edge robot developed by the ISR laboratory. The HermesBot is equipped with a comprehensive array of sensors that enable it to perceive its surroundings and navigate autonomously. These sensors include an RGBD Realsense D435 camera, which captures high-quality RGB and depth images, and a LIDAR Velodyne VLP-16, capable of generating a real-time 3D map of the robot's environment.
\begin{figure}[h!]
\centering
    \includegraphics[width=0.3\textwidth]{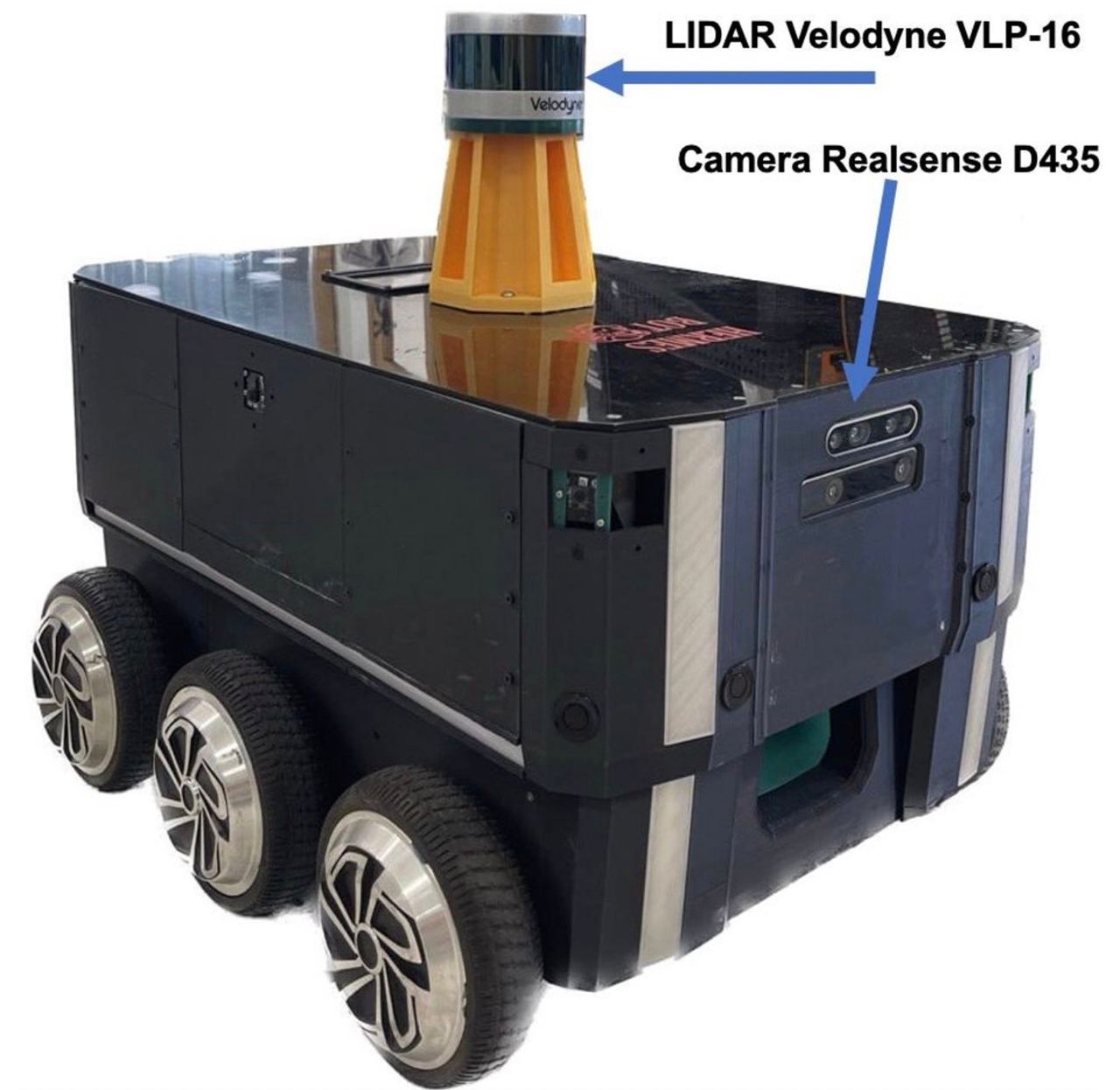}
    \caption{The HermesBot is a state-of-the-art robot, which equipped with advanced sensors for autonomous navigation and environment perception.} \label{fig:hermes}
\end{figure}
To perform the complex computations required for autonomous navigation and obstacle avoidance, the HermesBot is powered by two separate computers. The main computer is an Intel NUC with a Core i7 processor, which is responsible for high-level control and decision-making. The second computer is an NVIDIA Jetson Xavier, which is specifically designed for running deep learning algorithms efficiently. Together, these powerful computers allow the HermesBot to process large amounts of data in real-time and make intelligent decisions based on its environment. For neural networks training was used Nvidia RTX 3070 GPU and intel core i7 10700k. HermesBot is illustrated in Fig. \ref{fig:hermes}

\section{Experiments}
\subsection{Experiment 1, creating 3D reconstruction}
\begin{figure*}[h]
\centering
\includegraphics[width=0.9\textwidth]{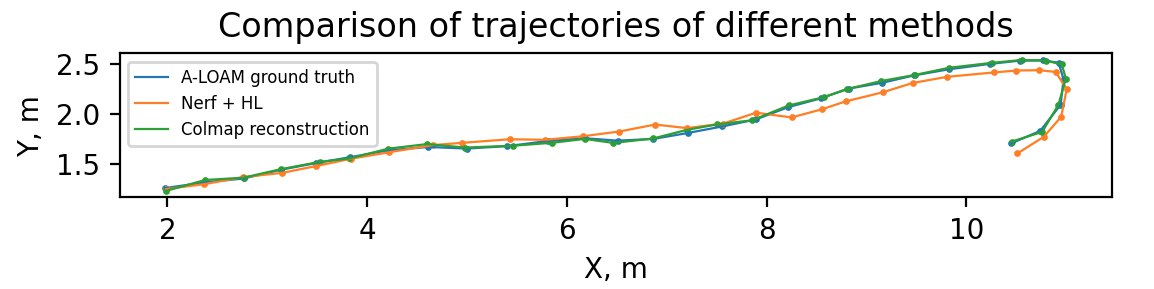}
\caption{Estimation of image positions with A-LOAM SLAM}
\label{fig:trajectory_plot}
\vspace{-1.5em}
\end{figure*} 

To evaluate the accuracy and effectiveness of the proposed approach for improving the SFM method, several experiments were conducted. First, a 3D reconstruction was created by collecting 400 images using a 12-megapixel camera with a ƒ/1.6 aperture and LIDAR, of which 200 were retained after removing blurred images. The NeRF reconstruction was performed using NerfStudio tool. 

\begin{figure}[h]
\centering
  \hspace*{\fill}
  \begin{subfigure}{0.22\textwidth}
    \includegraphics[width=\linewidth]{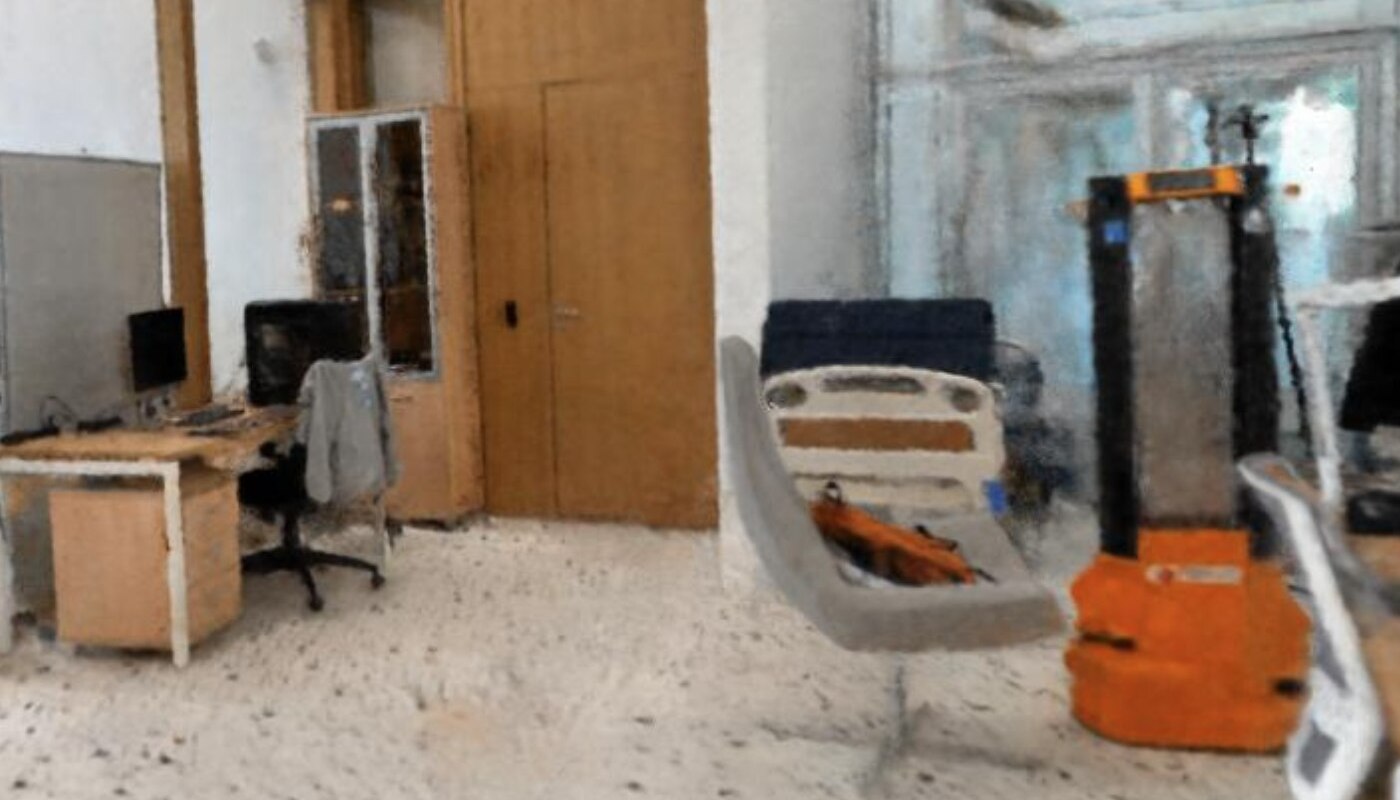}
    \caption{} \label{fig:continuous_1a}
  \end{subfigure}
  \hspace*{\fill}
  \begin{subfigure}{0.22\textwidth}
    \includegraphics[width=\linewidth]{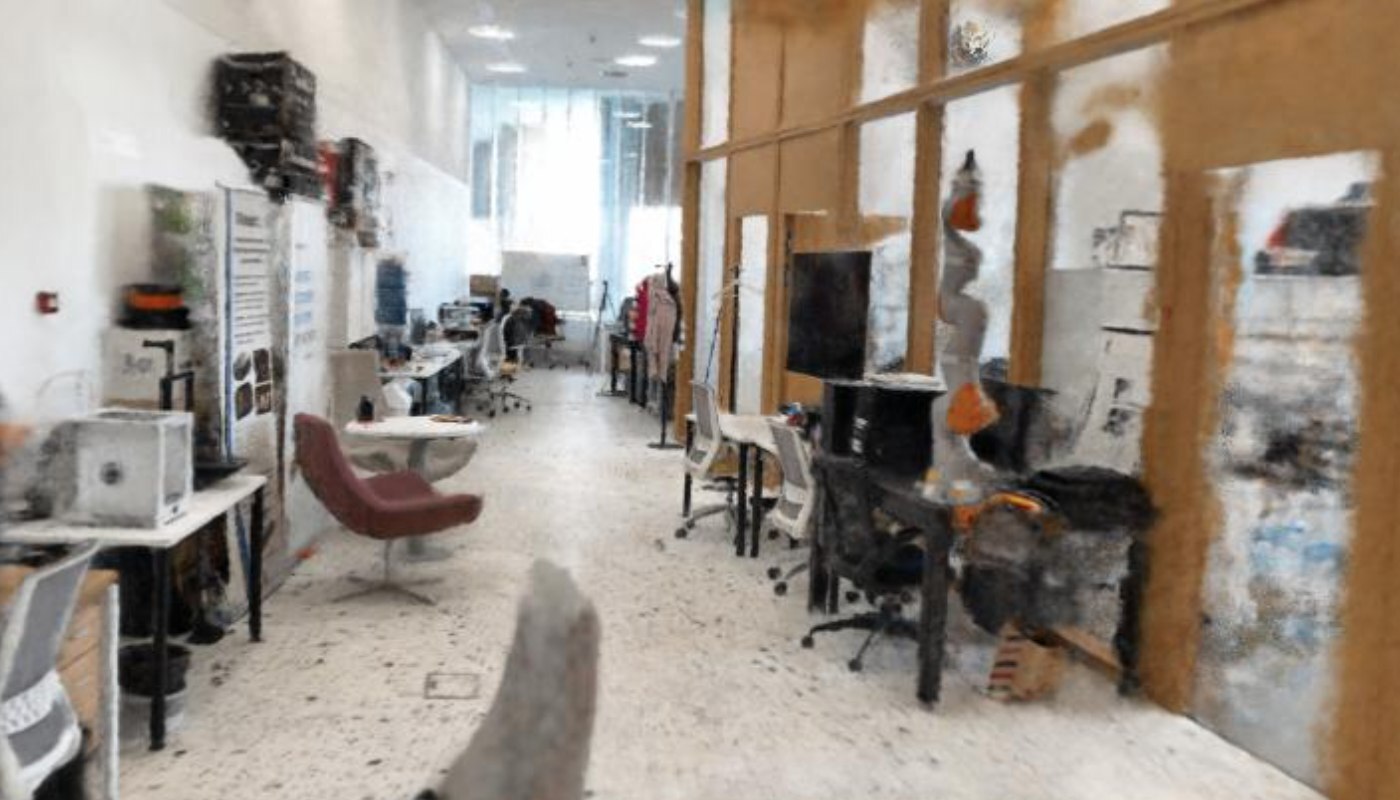}
    \caption{} \label{fig:continuous_1b}
  \end{subfigure}
  \hspace*{\fill}
  \\
    \hspace*{\fill}
    \begin{subfigure}{0.22\textwidth}
    \includegraphics[width=\linewidth]{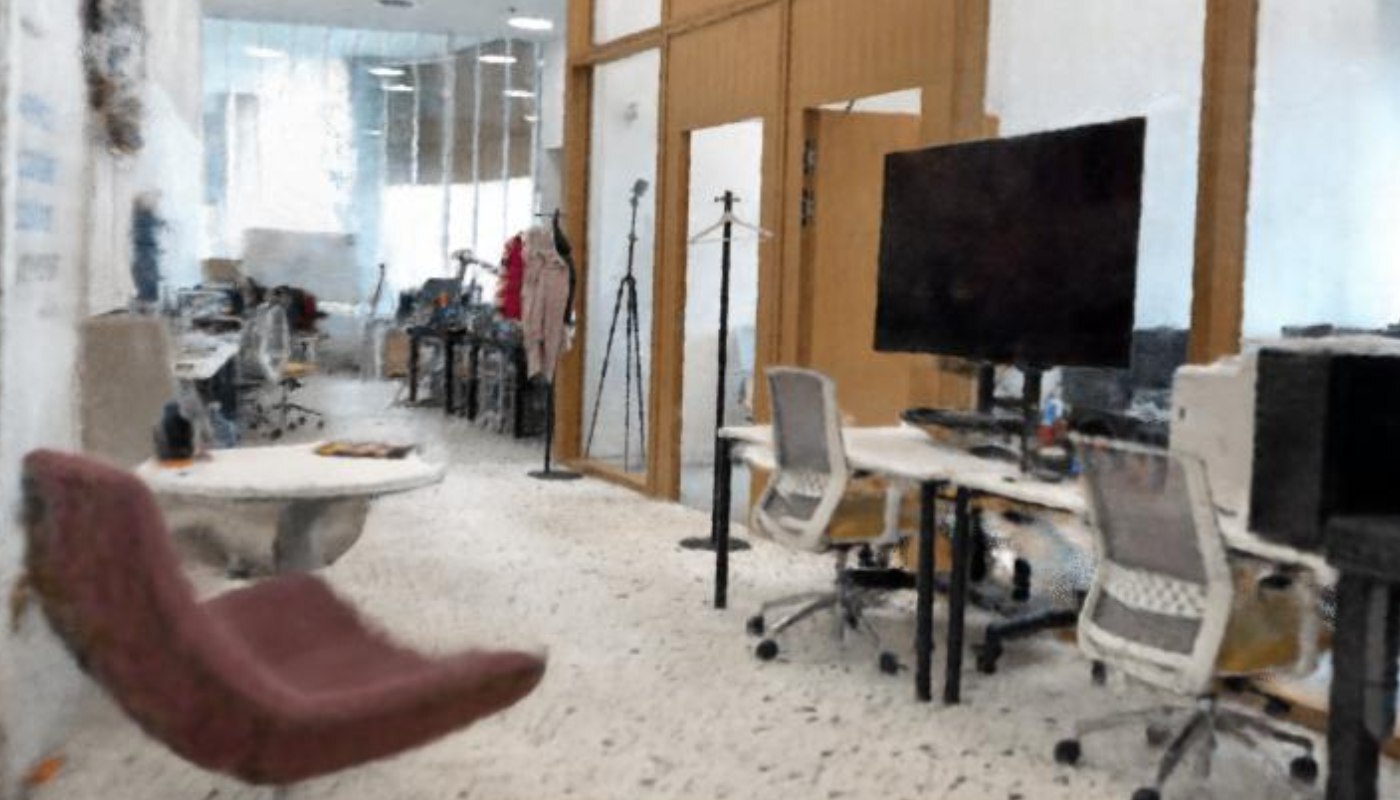}
    \caption{} \label{fig:continuous_1c}
  \end{subfigure}
  \hspace*{\fill}
    \begin{subfigure}{0.22\textwidth}
    \includegraphics[width=\linewidth]{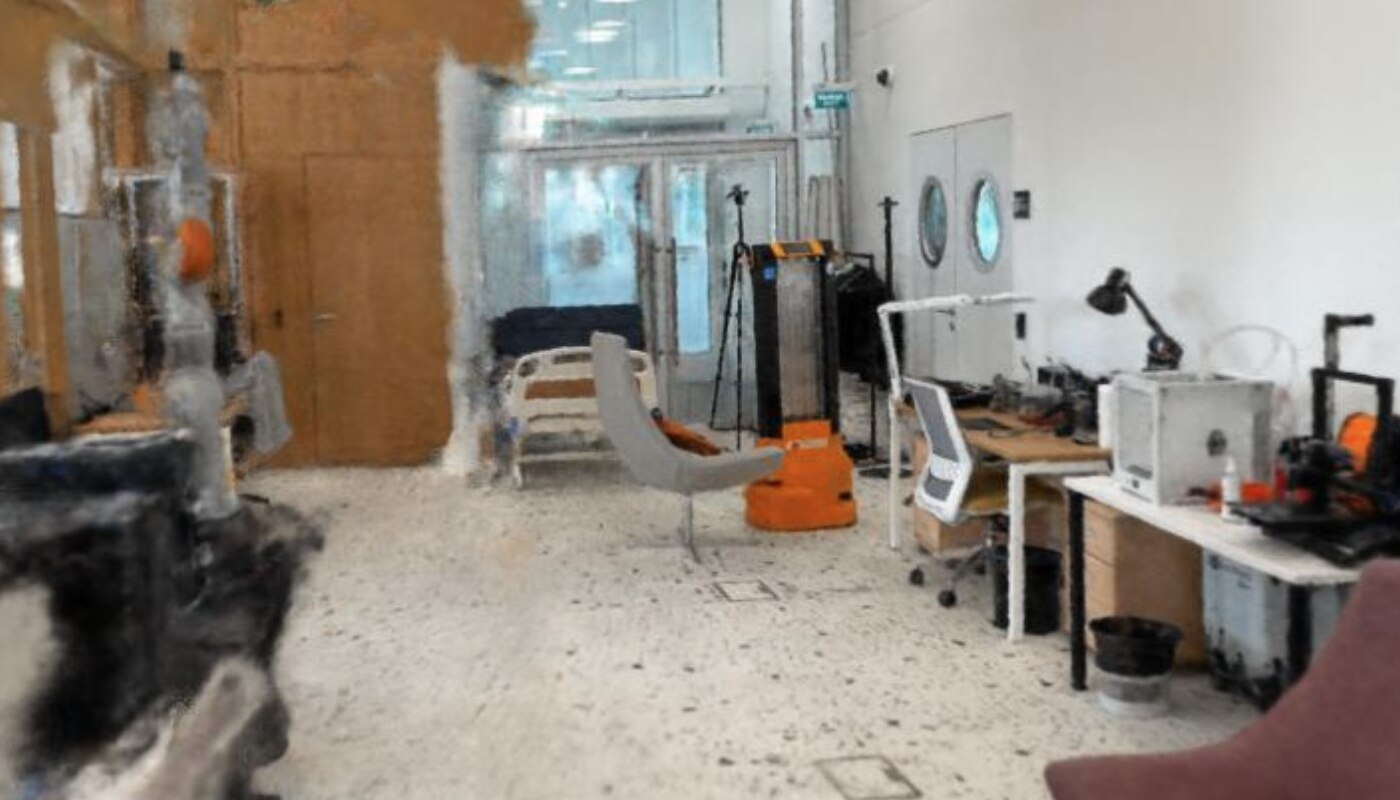}
    \caption{} \label{fig:continuous_1d}
  \end{subfigure}
  \hspace*{\fill}
\caption{The images rendered from 3D reconstruction based on nerfacto method}
\label{fig:images_from_nerf}
\vspace{-1.5em}
\end{figure}

To validate the reconstruction, 20 images were taken and the Peak Signal-to-Noise Ratio (PSNR) metric was used. The poses were estimated using LIDAR data, and the model was trained using the Instant NGP and Instant NGP boxed methods, which took only 5 minutes. Based on the PSNR and visual metrics, the NerfActo method was selected to increase reconstruction quality, resulting in high-quality images. The metrics comparison is provided in Table \ref{nerf_training_results}. Result is reconstrution that sutable for performing localization approach. Rendered images provided in Fig. \ref{fig:images_from_nerf}. The result size of NeRF map is 1200 m\textsuperscript{3}

\begin{table}[!hb]
\caption{NeRF reconstruction comparison}
\label{nerf_training_results}
\resizebox{0.47\textwidth}{!}{%
\begin{tabular}{{|l}*{2}{|c}|}
\hline
 \textbf{Method} &\textbf{Training time, min} &\textbf{PSNR, dB} \\
 \hline
    Instant-NGP & 6 & 12.4\\
    \hline
    Instant-NGP boxed & 5 & 14.1\\
    \hline
    Nerfacto & 30 & 18.4\\
 \hline

\end{tabular}
}
\end{table}


\subsection{Experiment 2, performing localization algorithm}
We ran our localization pipeline on a dataset recorded by HermesBot by launching the robot in the reconstruction area. To validate our method, data from LIDAR was recorded. The trajectory length was 10 meters. The algorithm was run on an Nvidia RTX 3070 GPU and Intel Core i7 10700k, and the speed of one iteration was 5 seconds. The camera resolution of the robot was 1280x720, and the algorithm used a parameter for the number of images needed to be rendered to determine the robot's position. The optimal parameter was found to be 2 images, which located 20 cm to the left and right of the quary image. For comparison, the COLMAP method was also used on training images for reconstruction, which took 6 hours to build the trajectory. Fig. \ref{fig:trajectory_plot} shows the trajectories obtained, including the ground truth obtained using the A-LOAM method. The accuracy comparison of the methods relative to the ground truth is provided in Table \ref{methods_values}.

\begin{table}[!h]
\caption{Localization experiment result metrics}
\resizebox{0.47\textwidth}{!}{%
\begin{tabular}{|p{2cm}|p{1.3cm}|p{1.3cm}|p{1.3cm}|}
\hline
\label{methods_values}
\textbf{Method} &\textbf{Trajectory Error, m} &\textbf{Rotation Error, rad} &\textbf{Map size, megabytes}\\ 
\hline
COLMAP & 0.022 & 0.012 & 400\\
\hline
SFM with NeRF & 0.068 & 0.07 & 160\\
 \hline
\end{tabular}
}
\end{table}

\section{Conclusion and Future Work}
We have presented a novel method for visual localization using NeRF (Neural Radiance Fields) reconstruction. Our pipeline eliminates the need to store a database of images and achieves accuracy comparable to that of NeRF. Unlike existing methods, our approach can operate in real-time. Moreover, our reconstruction technique provides additional information, such as object segmentation. Based on experiments, our method demonstrates promising results in terms of accuracy. Specifically, our method achieves an accuracy of 0.068 compared to the ground truth, with good translation and extension performance.Our proposed method offers several advantages over existing techniques, including a 50\% reduction in storage requirements and real-time performance. By leveraging the power of NeRF reconstruction, we can obtain high-quality reconstructions that provide valuable additional information in 1200 m\textsuperscript{3} volume. Our results demonstrate the potential of our method for applications in robotics, autonomous vehicles, and augmented reality. Overall, our approach represents a significant step forward in visual localization and has the potential to impact a wide range of fields.

The future work is to apply our method to larger indoor and outdoor spaces \cite{protasov2021cnn}, includes real-world conditions on various setups, for example, outdoor plant inspection robots \cite{karpyshev2021autonomous}, indoor ground robots for charging \cite{okunevich2021deltacharger} and shopping rooms \cite{petrovsky2020customer}, and UAVs \cite{kalinov2021impedance, kalinov2019high, kalinov2020warevision, kalinov2021warevr, yatskin2017principles}, as well as examine the applicability of the approach to more sophisticated systems, e.g., modular two-wheeled rovers \cite{petrovsky2022two} and VR reconstruction \cite{ponomareva2021grasplook}. We also plan to conduct more experiments with the number of images needed for accurate localization. Additionally, we aim to implement a mechanism to support a certain size of the database and remove images with few features, thereby reducing the algorithm's processing time.

\bibliographystyle{ieeetr}
\bibliography{bib.bib}

\begin{thebibliography}{10}

\bibitem{chen2020survey}
C.~Chen, B.~Wang, C.~X. Lu, N.~Trigoni, and A.~Markham, ``A survey on deep
  learning for localization and mapping: Towards the age of spatial machine
  intelligence,'' {\em arXiv preprint arXiv:2006.12567}, 2020.

\bibitem{kalinov2020warevision}
I.~Kalinov, A.~Petrovsky, V.~Ilin, E.~Pristanskiy, M.~Kurenkov, V.~Ramzhaev,
  I.~Idrisov, and D.~Tsetserukou, ``Warevision: Cnn barcode detection-based uav
  trajectory optimization for autonomous warehouse stocktaking,'' {\em IEEE
  Robotics and Automation Letters}, vol.~5, no.~4, pp.~6647--6653, 2020.

\bibitem{perminov2021ultrabot}
S.~Perminov, N.~Mikhailovskiy, A.~Sedunin, I.~Okunevich, I.~Kalinov,
  M.~Kurenkov, and D.~Tsetserukou, ``Ultrabot: Autonomous mobile robot for
  indoor uv-c disinfection,'' in {\em 2021 IEEE 17th International Conference
  on Automation Science and Engineering (CASE)}, pp.~2147--2152, IEEE, 2021.

\bibitem{karpyshev2022mucaslam}
P.~Karpyshev, E.~Kruzhkov, E.~Yudin, A.~Savinykh, A.~Potapov, M.~Kurenkov,
  A.~Kolomeytsev, I.~Kalinov, and D.~Tsetserukou, ``Mucaslam: Cnn-based frame
  quality assessment for mobile robot with omnidirectional visual slam,'' in
  {\em 2022 IEEE 18th International Conference on Automation Science and
  Engineering (CASE)}, pp.~368--373, IEEE, 2022.

\bibitem{mikhailovskiy2021ultrabot}
N.~Mikhailovskiy, A.~Sedunin, S.~Perminov, I.~Kalinov, and D.~Tsetserukou,
  ``Ultrabot: Autonomous mobile robot for indoor uv-c disinfection with
  non-trivial shape of disinfection zone,'' in {\em 2021 26th IEEE
  International Conference on Emerging Technologies and Factory Automation
  (ETFA)}, pp.~1--7, IEEE, 2021.

\bibitem{schonberger2016structure}
J.~L. Schonberger and J.-M. Frahm, ``Structure-from-motion revisited,'' in {\em
  Proceedings of the IEEE conference on computer vision and pattern
  recognition}, pp.~4104--4113, 2016.

\bibitem{van2018efficient}
D.~Van~Opdenbosch, T.~Aykut, N.~Alt, and E.~Steinbach, ``Efficient map
  compression for collaborative visual slam,'' in {\em 2018 IEEE winter
  conference on applications of computer vision (WACV)}, pp.~992--1000, IEEE,
  2018.

\bibitem{mera2020efficient}
M.~Mera-Trujillo, B.~Smith, and V.~Fragoso, ``Efficient scene compression for
  visual-based localization,'' in {\em 2020 International Conference on 3D
  Vision (3DV)}, pp.~1--10, IEEE, 2020.

\bibitem{camposeco2019hybrid}
F.~Camposeco, A.~Cohen, M.~Pollefeys, and T.~Sattler, ``Hybrid scene
  compression for visual localization,'' in {\em Proceedings of the IEEE/CVF
  Conference on Computer Vision and Pattern Recognition}, pp.~7653--7662, 2019.

\bibitem{ullman1979interpretation}
S.~Ullman, ``The interpretation of structure from motion,'' {\em Proceedings of
  the Royal Society of London. Series B. Biological Sciences}, vol.~203,
  no.~1153, pp.~405--426, 1979.

\bibitem{sattler2017large}
T.~Sattler, A.~Torii, J.~Sivic, M.~Pollefeys, H.~Taira, M.~Okutomi, and
  T.~Pajdla, ``Are large-scale 3d models really necessary for accurate visual
  localization?,'' in {\em Proceedings of the IEEE Conference on Computer
  Vision and Pattern Recognition}, pp.~1637--1646, 2017.

\bibitem{sarlin2019coarse}
P.-E. Sarlin, C.~Cadena, R.~Siegwart, and M.~Dymczyk, ``From coarse to fine:
  Robust hierarchical localization at large scale,'' in {\em Proceedings of the
  IEEE/CVF Conference on Computer Vision and Pattern Recognition},
  pp.~12716--12725, 2019.

\bibitem{rublee2011orb}
E.~Rublee, V.~Rabaud, K.~Konolige, and G.~Bradski, ``Orb: An efficient
  alternative to sift or surf,'' in {\em 2011 International conference on
  computer vision}, pp.~2564--2571, Ieee, 2011.

\bibitem{bay2006surf}
H.~Bay, T.~Tuytelaars, and L.~V. Gool, ``Surf: Speeded up robust features,'' in
  {\em European conference on computer vision}, pp.~404--417, Springer, 2006.

\bibitem{ng2003sift}
P.~C. Ng and S.~Henikoff, ``Sift: Predicting amino acid changes that affect
  protein function,'' {\em Nucleic acids research}, vol.~31, no.~13,
  pp.~3812--3814, 2003.

\bibitem{arandjelovic2016netvlad}
R.~Arandjelovic, P.~Gronat, A.~Torii, T.~Pajdla, and J.~Sivic, ``Netvlad: Cnn
  architecture for weakly supervised place recognition,'' in {\em Proceedings
  of the IEEE conference on computer vision and pattern recognition},
  pp.~5297--5307, 2016.

\bibitem{detone2018superpoint}
D.~DeTone, T.~Malisiewicz, and A.~Rabinovich, ``Superpoint: Self-supervised
  interest point detection and description,'' in {\em Proceedings of the IEEE
  conference on computer vision and pattern recognition workshops},
  pp.~224--236, 2018.

\bibitem{sarlin2021back}
P.-E. Sarlin, A.~Unagar, M.~Larsson, H.~Germain, C.~Toft, V.~Larsson,
  M.~Pollefeys, V.~Lepetit, L.~Hammarstrand, F.~Kahl, {\em et~al.}, ``Back to
  the feature: Learning robust camera localization from pixels to pose,'' in
  {\em Proceedings of the IEEE/CVF conference on computer vision and pattern
  recognition}, pp.~3247--3257, 2021.

\bibitem{yudin2022cloudvision}
E.~Yudin, P.~Karpyshev, M.~Kurenkov, A.~Savinykh, A.~Potapov, E.~Kruzhkov, and
  D.~Tsetserukou, ``Cloudvision: Dnn-based visual localization of autonomous
  robots using prebuilt lidar point cloud,'' {\em arXiv preprint
  arXiv:2209.01605}, 2022.

\bibitem{mildenhall2021nerf}
B.~Mildenhall, P.~P. Srinivasan, M.~Tancik, J.~T. Barron, R.~Ramamoorthi, and
  R.~Ng, ``Nerf: Representing scenes as neural radiance fields for view
  synthesis,'' {\em Communications of the ACM}, vol.~65, no.~1, pp.~99--106,
  2021.

\bibitem{barron2021mip}
J.~T. Barron, B.~Mildenhall, M.~Tancik, P.~Hedman, R.~Martin-Brualla, and P.~P.
  Srinivasan, ``Mip-nerf: A multiscale representation for anti-aliasing neural
  radiance fields,'' in {\em Proceedings of the IEEE/CVF International
  Conference on Computer Vision}, pp.~5855--5864, 2021.

\bibitem{barron2022mip}
J.~T. Barron, B.~Mildenhall, D.~Verbin, P.~P. Srinivasan, and P.~Hedman,
  ``Mip-nerf 360: Unbounded anti-aliased neural radiance fields,'' in {\em
  Proceedings of the IEEE/CVF Conference on Computer Vision and Pattern
  Recognition}, pp.~5470--5479, 2022.

\bibitem{zhang2020nerf++}
K.~Zhang, G.~Riegler, N.~Snavely, and V.~Koltun, ``Nerf++: Analyzing and
  improving neural radiance fields,'' {\em arXiv preprint arXiv:2010.07492},
  2020.

\bibitem{sitzmann2020implicit}
V.~Sitzmann, J.~Martel, A.~Bergman, D.~Lindell, and G.~Wetzstein, ``Implicit
  neural representations with periodic activation functions,'' {\em Advances in
  Neural Information Processing Systems}, vol.~33, pp.~7462--7473, 2020.

\bibitem{muller2022instant}
T.~M{\"u}ller, A.~Evans, C.~Schied, and A.~Keller, ``Instant neural graphics
  primitives with a multiresolution hash encoding,'' {\em ACM Transactions on
  Graphics (ToG)}, vol.~41, no.~4, pp.~1--15, 2022.

\bibitem{tancik2023nerfstudio}
M.~Tancik, E.~Weber, E.~Ng, R.~Li, B.~Yi, J.~Kerr, T.~Wang, A.~Kristoffersen,
  J.~Austin, K.~Salahi, {\em et~al.}, ``Nerfstudio: A modular framework for
  neural radiance field development,'' {\em arXiv preprint arXiv:2302.04264},
  2023.

\bibitem{maggio2022loc}
D.~Maggio, M.~Abate, J.~Shi, C.~Mario, and L.~Carlone, ``Loc-nerf: Monte carlo
  localization using neural radiance fields,'' {\em arXiv preprint
  arXiv:2209.09050}, 2022.

\bibitem{zhang2014loam}
J.~Zhang and S.~Singh, ``Loam: Lidar odometry and mapping in real-time.,'' in
  {\em Robotics: Science and Systems}, vol.~2, pp.~1--9, Berkeley, CA, 2014.

\bibitem{wang2021f}
H.~Wang, C.~Wang, C.-L. Chen, and L.~Xie, ``F-loam: Fast lidar odometry and
  mapping,'' in {\em 2021 IEEE/RSJ International Conference on Intelligent
  Robots and Systems (IROS)}, pp.~4390--4396, IEEE, 2021.

\bibitem{protasov2021cnn}
S.~Protasov, P.~Karpyshev, I.~Kalinov, P.~Kopanev, N.~Mikhailovskiy,
  A.~Sedunin, and D.~Tsetserukou, ``Cnn-based omnidirectional object detection
  for hermesbot autonomous delivery robot with preliminary frame
  classification,'' in {\em 2021 20th International Conference on Advanced
  Robotics (ICAR)}, pp.~517--522, IEEE, 2021.

\bibitem{karpyshev2021autonomous}
P.~Karpyshev, V.~Ilin, I.~Kalinov, A.~Petrovsky, and D.~Tsetserukou,
  ``Autonomous mobile robot for apple plant disease detection based on cnn and
  multi-spectral vision system,'' in {\em 2021 IEEE/SICE international
  symposium on system integration (SII)}, pp.~157--162, IEEE, 2021.

\bibitem{okunevich2021deltacharger}
I.~Okunevich, D.~Trinitatova, P.~Kopanev, and D.~Tsetserukou, ``Deltacharger:
  Charging robot with inverted delta mechanism and cnn-driven high fidelity
  tactile perception for precise 3d positioning,'' {\em IEEE Robotics and
  Automation Letters}, vol.~6, no.~4, pp.~7604--7610, 2021.

\bibitem{petrovsky2020customer}
A.~Petrovsky, I.~Kalinov, P.~Karpyshev, M.~Kurenkov, V.~Ramzhaev, V.~Ilin, and
  D.~Tsetserukou, ``Customer behavior analytics using an autonomous
  robotics-based system,'' in {\em 2020 16th International Conference on
  Control, Automation, Robotics and Vision (ICARCV)}, pp.~327--332, IEEE, 2020.

\bibitem{kalinov2021impedance}
I.~Kalinov, A.~Petrovsky, R.~Agishev, P.~Karpyshev, and D.~Tsetserukou,
  ``Impedance-based control for soft uav landing on a ground robot in
  heterogeneous robotic system,'' in {\em 2021 International Conference on
  Unmanned Aircraft Systems (ICUAS)}, pp.~1653--1658, IEEE, 2021.

\bibitem{kalinov2019high}
I.~Kalinov, E.~Safronov, R.~Agishev, M.~Kurenkov, and D.~Tsetserukou,
  ``High-precision uav localization system for landing on a mobile
  collaborative robot based on an ir marker pattern recognition,'' in {\em 2019
  IEEE 89th Vehicular Technology Conference (VTC2019-Spring)}, pp.~1--6, IEEE,
  2019.

\bibitem{kalinov2021warevr}
I.~Kalinov, D.~Trinitatova, and D.~Tsetserukou, ``Warevr: Virtual reality
  interface for supervision of autonomous robotic system aimed at warehouse
  stocktaking,'' in {\em 2021 ieee international conference on systems, man,
  and cybernetics (smc)}, pp.~2139--2145, IEEE, 2021.

\bibitem{yatskin2017principles}
D.~Yatskin and I.~Kalinov, ``Principles of solving the space monitoring problem
  by multirotors swarm,'' in {\em 2017 IVth International Conference on
  Engineering and Telecommunication (EnT)}, pp.~47--50, IEEE, 2017.

\bibitem{petrovsky2022two}
A.~Petrovsky, I.~Kalinov, P.~Karpyshev, D.~Tsetserukou, A.~Ivanov, and
  A.~Golkar, ``The two-wheeled robotic swarm concept for mars exploration,''
  {\em Acta Astronautica}, vol.~194, pp.~1--8, 2022.

\bibitem{ponomareva2021grasplook}
P.~Ponomareva, D.~Trinitatova, A.~Fedoseev, I.~Kalinov, and D.~Tsetserukou,
  ``Grasplook: a vr-based telemanipulation system with r-cnn-driven
  augmentation of virtual environment,'' in {\em 2021 20th International
  Conference on Advanced Robotics (ICAR)}, pp.~166--171, IEEE, 2021.

\end{thebibliography}

\end{document}